\documentclass[review]{elsarticle}

\usepackage{lineno,hyperref}
\usepackage{caption}
\usepackage{mathtools}
\usepackage{amsfonts, amsmath, amssymb, amsthm}
\usepackage{tabularx,ragged2e}
\usepackage{array}
\usepackage{booktabs}

\usepackage{latexsym}
\usepackage{bm,array}
\usepackage{calc}

\usepackage[table,xcdraw]{xcolor}
\usepackage{pdfpages}
\usepackage{graphicx}

\journal{tbc}

\begin{document}

\begin{frontmatter}

\title{Semantic Segmentation of Human Thigh Quadriceps Muscle in Magnetic Resonance Images}

\author{E. Ahmad\fnref{myfootnote}\corref{equalcontributingauthor}}
\author{M. Goyal\fnref{myfootnote}\corref{equalcontributingauthor}}
\author{J. S. McPhee\fnref{myfootnote2}}
\author{H. Degens\fnref{myfootnote2}}
\author{and M. H. Yap\fnref{myfootnote}}
\address{Manchester Metropolitan University, Manchester, United Kingdom}
\fntext[myfootnote]{School of Computing, Mathematics and Digital Technology.}
\fntext[myfootnote2]{School of Healthcare Science.}
\cortext[equalcontributingauthor]{Authors Contributed Equally}




\begin{abstract}
This paper presents an end-to-end solution for MRI thigh quadriceps segmentation. This is the first attempt that deep learning methods are used for the MRI thigh segmentation task. We use the state-of-the-art Fully Convolutional Networks with transfer learning approach for the semantic segmentation of regions of interest in MRI thigh scans. To further improve the performance of the segmentation, we propose a post-processing technique using basic image processing methods. With our proposed method, we have established a new benchmark for MRI thigh quadriceps segmentation with mean \textit{Jaccard Similarity Index} of 0.9502 and processing time of 0.117 second per image.
\end{abstract}


\end{frontmatter}


\section{Introduction}

\noindent There are over 600 skeletal muscles in the human body which play important roles in: 1) thermoregulation by generating heat as a product of their contraction and metabolism; 2) whole-body energy balance by utilizing and storing fatty acids, glucose and amino acids consumed as part of the diet to provide energy for movement and other cellular processes; and 3) paracrine and endocrine functions as they release growth hormones and other factors into circulation \cite{blakey1992muscle}. However, the most important functions of muscles are to provide support for posture and to make movement possible by applying force while shortening to bones and joints.

The muscles of the arms and legs principally allow us to interact with and move around the environment. Of all the limb muscles, those located in the thigh are amongst the largest and most powerful. The thigh is divided into three main compartments: 1) the quadriceps muscle group on the anterior aspect; 2) the adductor muscles that bring the thigh toward the mid-line and rotate it (on the medial section); and 3) the hamstrings on the posterior aspect.

Skeletal muscles are highly adaptable to their habitual use: regular exercise will induce muscles growth, while sedentary living or other types of disuse leads to muscle atrophy. Indeed, any disease or other long-term condition causing low muscle mass and strength can result in mobility impairments. The normal ageing process is one such condition. During ageing, physical activity levels decrease \cite{sallis2000age} and the consequent disuse alongside other biological processes such as hormonal and/or other endocrine changes contribute to muscle wasting \cite{degens2010role}. 

Another study has revealed that leg muscles of older people begin to accumulate fatty adipose tissue. There is very little information currently available showing how adipose tissue accumulates in leg muscles. Quantification of the changes in adipose tissue with ageing coupled with more detailed analysis of muscle biopsy samples could help to identify the regulatory mechanisms leading to loss of muscle mass and strength in older age because fatty adipose tissue can emit proteins that promote the breakdown of surrounding tissue \cite{picard2005molecular}. Hence, The study of thigh muscles is a priority due to their importance in locomotion, mobility and metabolism and the extent of their deterioration in ageing and disease.  

Artificial Intelligence is showing improvements in the autonomous and unsupervised learning systems, which in turn, revolutionize many industries and bring significant shifts in society through developments in healthcare analysis and treatment \cite{liang2014deep, granter2017straw, goyal2017dfunet}. Technological advances in medical imaging over the past three decades have led to greater accessibility to advanced imaging techniques in clinical practice and research settings. Such technologies include magnetic resonance imaging (MRI), dual-energy X-ray absorptiometry, ultrasonography and computed tomography. They are widely used by clinicians to scan body segments to diagnose injury or disease \cite{yap2010processed, yap2008generic}. These imaging techniques are also used by researchers interested in studies of human anatomy and physiology in health and disease across the lifespan. These techniques produce precise, high-quality distinctions between different tissue and cell types.

In computer vision analysis and medical imaging, semantic segmentation is an attempt to partition meaningful parts that belong to the same object class of an image together and is mostly applied in object detection and localization \cite{thoma2016survey,shelhamer2017fully}. In this work, five state-of-the-art semantic segmentation deep learning architectures for regions of interest (ROI) in MRI of thigh muscles are reviewed.

The significant contributions of this work are:
\begin{enumerate}
	\item To provide a review of current state-of-the-art image processing, traditional machine learning and deep learning approaches for segmentation of MRI thigh muscles.
	\item We present the largest dataset of MRI thigh muscles with annotated ROI that consists of quadriceps, bone and marrow as ground truths from the experts.
	\item To date, this is the first time deep learning methods are used to segment the ROI from the MRI of thigh muscles and results are compared with the traditional methods.
	\item We propose a post-processing technique which helps in further refinement of segmentation of ROI in MRI of thigh muscles.
\end{enumerate} 

\section{Related Work}

\noindent From 2002 until 2015, twelve automated segmentation attempts on the MRI thigh muscles have been recorded \cite{barra2002segmentation, mattei2006segmentation, kang2007automatic, kang2007tissue, wang2007practical, urricelqui2009automatic, positano2009accurate, jiang2010research, makrogiannis2012automated, purushwalkam2013automatic, valentinitsch2013automated, orgiu2015automatic}), where automation process was done by incorporating either one or combination of these techniques:  thresholding (intensity based or histogram modelling); classification (fuzzy c-means (FCM) being the popular approach or k-means); active contour; and/or region growing. These methods focusing on segmenting muscles, marrow, femur, subcutaneous adipose tissue and/or intermuscular adipose tissue as an individual component (or group), whereafter the implementation of suitable pre-processing algorithms (to remove/reduce noise or improve pixel's intensity), this individual component (or group) can be straightforwardly segmented. 

The segmentation accuracies across all methods above are considered exceptional in general, with all methods achieved average accuracy of greater than 85\% and the segmentation results can be seen statistically improved over time (2006/07 \cite{kang2007automatic, kang2007tissue} - 95.73\%, by FCM; 2009 \cite{urricelqui2009automatic} - about 96\%, by adaptive thresholding and histogram modelling; 2013 \cite{valentinitsch2013automated} - about 97\%, by \textit{k}-means clustering; and 2015 \cite{orgiu2015automatic} - 96.8\%, by snake active contour). Considering the application (or combination of applications) of such basic techniques, the system's average processing time across all methods are also recorded to be improved over time (2009 \cite{positano2009accurate} - 52 sec per image, by FCM and active contour; 2013 \cite{purushwalkam2013automatic} - about 5.21 sec per image, by region growing and 3D intensity map; and 2015 \cite{valentinitsch2013automated} - 0.25 sec per image, by \textit{k}-means clustering). Two major reasons that understandably contribute to the better mean average processing time are superior computer technology and hardware available on the market and the optimisation of algorithms and processing platforms from the developer. Regardless, this simple, precise and fast automation process may also assist significantly in quantification analysis for cross-sectional area or volumetric of MRI of the thigh, to a certain extent.

In 2012, an advanced discrete optimization solution by a graph-based Random Walker (RW) has been proposed by \textit{Baudin et al.} \cite{baudin2012automatic}, where a graphical model that capable to automatically determine appropriate seed positions with respect to different muscle classes has been introduced. The process was established by combining frameworks of seeds sampling and graph edges; and Markov Random Fields formulation that calculate the cost function form, unary potential, geodesic distance potential, and relative orientation potential of muscle structure and position. Authors reported that small muscles (within the thigh region) are prone to segmentation errors compared to the large ones and as a result, recommending the use of deformable registration approaches to alleviate the bias introduced from the rigid registration step, for future work. Next, in \cite{baudin2012manifold}, the same authors worked on the introduction of linear sub-spaces constraints within RW segmentation framework, where the novel knowledge or prior-based quadriceps segmentation was proposed, contributing to a slightly improved in segmentation output accuracy, compared to \cite{baudin2012automatic}. The technique was developed based on RW formulation, where the concatenation of all the node's (or seed's) probability is proliferated by block diagonal matrix (depending on the quantity of the label); combined with prior knowledge to the respective RW formulation above; and the design of low dimensional affine of implicit space within the spatial matrix. Results were presented in 2D MRI cross-sectional area, with an average segmentation accuracy (in Dice Similarity Coefficient (DSC)) of $0.84 \pm 0.08$ and computational time of approximately 15 minutes per segmentation. Another effort from \textit{Baudin et al.} including an extension for the segmentation of whole muscles of thigh MRI \cite{baudin2012prior}, again by integration of similar RW graph-based with prior information frameworks as above, where statistical shape atlas was employed to represent prior knowledge. Apart from this frameworks, a balancing parameter was introduced to the formulation of nodal's total energy; with an addition of confidence map to adjust the influence of contour's model. This configuration leads to an efficient iterative linear optimization and further enhanced the quality of segmentation accuracy, yielding an average DSC of $0.86 \pm 0.07$ and processing time of around 5 minutes per segmentation. However, errors in muscle segmentation are still occurred, especially to small muscle components (e.g. Gracilis muscle and tensor fasciae latae), due to large registration errors on the same muscle assumption.

Another fully automatic segmentation method for thigh MRI was demonstrated by \textit{Andrews et al.} \cite{andrews2011probabilistic}. This technique provides a good accuracy for individual muscle segmentation of the thigh, by using energy minimizing probabilistic segmentation that indicates area of ambiguity. The automatic segmentation made possible with the frameworks that comprise of a probabilistic segmentation function representation of isometric log-ratio (ILR) transform; the shape space (relative locations of muscles) by principal component analysis (PCA); image alignment by using femur bone coordination as an anchor of the vector estimation; and energy construction strictly based on energy function over the shape space of an image. The results were presented in 3D and the method capable to segment 11 muscles within the thigh automatically and achieved a mean DSC of $0.92 \pm 0.03$ across all images, but with undisclosed processing time per segmentation/image. A recent publication by \textit{Andrews and Hamarneh} \cite{andrews2015generalized} presents comprehensive incorporation of probabilistic shape representation consists of adjacent object information and prior anatomical volume, called generalized log-ratio (GLR). The main benefit of GLR application is that it may produce a probabilistic segmentation that can be used to generate uncertainty information to aid subsequent analysis of different muscle components on thigh MRI. In other words, the GLR transformation can be designed to ensure statistical shape models capture variability in smaller structures properly. The method is designed based on four major steps. Firstly, GLR representation in the context of thigh muscle segmentation to encode details such as muscle size, adjacency information, and to train a statistical shape model over the space of GLR representations. Secondly, the introduction of pre-segmenting images technique for all components involve (fat, muscles and bone). The results are mainly used to perform alignment computation. Thirdly, random forest classifier training to assist in detecting intermuscular boundary locations. And finally, results from all steps above were integrated into a globally minimized convex energy functional by the mean of primal-dual method to generate a probabilistic segmentation. As the system was mainly designed to tackle issues related to segmentation accuracy, the major drawback is that the average segmentation time took $50 \pm 4.3$ minutes per image to run. The mean DSC by using this method is $0.808 \pm 0.074$, which is quite promising, considering the procedure and analysis were done to all muscles, individually, on MRI of the thigh.

In 1992, \textit{Morrison et al.} \cite{morrison1992probabilistic} made an attempt on MRI thigh segmentation by using neural network. The network model integrates a probabilistic neural network to facilitate the generation of probability estimates at each pixel for use in an iterative segmentation process. This preliminary study on MRI thigh segmentation by using one of the simplest forms of neural network reported that the segmentation result is unaffected by the quality of the training dataset. Today, the concept of neural network has been revolutionised and diversely applied, including in segmentation of medical imaging. 

Recently, deep learning has provided various end-to-end solutions for the semantic segmentation of abnormalities such as breast cancer, brain tumour, skin lesions, foot ulcers in various image modalities of medical imaging \cite{goyal2017fully, yap2017automated, goyal2017multi}. At the time of writing, there is no other study has been made in the segmentation of MR images of human thigh muscles with deep learning methods. Current state-of-the-art methods, especially image processing based techniques, are not robust, due to their nature of reliance on specific regulators and rules, with certain assumptions. In contrast to conventional machine learning, deep learning methods do not require such strong assumptions and have demonstrated superiority in region and object segmentation and classification, which suggests that the robust fully automated MRI segmentation of human thigh muscles may be achieved, by adopting such approach. Moreover, the segmentation output performance from most recent applications of network models in MRI of different domains \cite{liao2013representation, baumgartner2017exploration, trebeschi2017deep}, in association with image processing technique \cite{avendi2016combined} and in conjunction with the utilization of better and upgradable graphics processing unit (GPU), is remarkably accurate and fast.

\section{Methods}
\noindent This section discusses the MRI datasets of thigh muscles, the preparation of ground truth annotation and the proposed methods for experimentation of semantic segmentation.

\subsection{MRI Datasets of Thigh Muscles}

\begin{figure}[!h] 
\centering
\includegraphics[width=\textwidth]{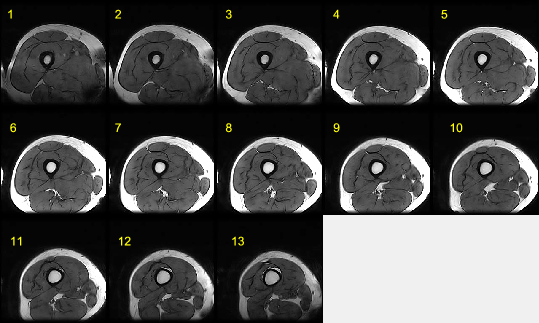}
\caption{13 sequential scans per individual subject.}
\label{Fig5}
\end{figure}

\begin{figure}[!h] 
\centering
\includegraphics[width=\textwidth]{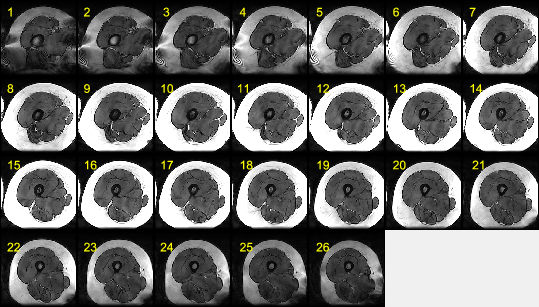}
\caption{26 sequential scans per individual subject}
\label{Fig6}
\end{figure}

\noindent All MRI scans of the thigh were collected from men and women aged 18-90 years \cite{ahmad2014enhancement} by using the same T1-weighted Turbo 3D sequences from a 0.2-Tesla MRI scanner (Esaote G-scan; Italy). All scans consisted of serial transverse-plane slices: 1) each with 6.3-mm thickness and 0-mm inter-slice gap; 2) image matrix of 256 x 256 (in pixels); and 3) assembled in either 13 or 26 sequential scans per individual subject. Figures \ref{Fig5} and \ref{Fig6} demonstrate both standard types of scan (per individual subject) with 13 and 26 scans, respectively. 

In this work, we created three types of MRI thigh datasets. The first dataset is labelled as mid-scan (MD), where a single middle-scanned MRI is selected from each dataset of 110 different subjects, i.e. MRI scan number 7 of individual subject with 13 scans (Figure \ref{Fig5}) and MRI scan number 13 of individual subject with 26 scans (Figure \ref{Fig6}). The second dataset is labelled as whole-scan (WD), which is the combination of all scans from both 13 and 26 scans datasets of 50 individual subjects to form a WD dataset with 1000 MR images. The third dataset is labelled as all-scan (AD), which is a combination of both MD and WD datasets.

Since, most of the pre-trained deep learning frameworks were trained on the three channel images such as RGB format, all of the original 1000 grayscale MRI are converted to the images of 3-channel colour profile of sRGB IEC61966-2.1. The binary ground truth masks which are delineated by domain experts to represent ROI are converted into the Pascal VOC format (labelled as 8-bit single-channel paletted images) \cite{shelhamer2017fully}.

In this Pascal VOC format, 0 maps are indexed to black pixels corresponding to the background and 1 (in red) denotes the ROI (consists of quadriceps, femur and marrow). This procedure is illustrated in Figure \ref{Fig7} below.

\begin{figure}[!h] 
\centering
\includegraphics[width=3.5in]{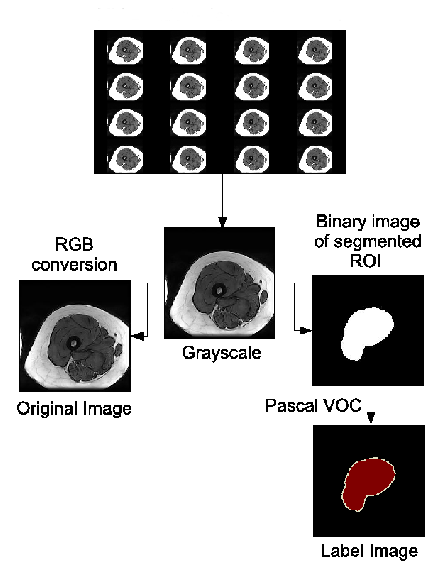}
\caption{Preparation of MRI Datasets.}
\label{Fig7}
\end{figure}

\subsection{Deep Learning Architectures for Semantic Segmentation}

\noindent This section provides the brief description of deep learning architectures used for the semantic segmentation of ROI in the MRI thigh muscles.  

\paragraph{\textbf{FCN-AlexNet}}

\textit{FCN-AlexNet} is a customized FCN model based on AlexNet architecture \cite{krizhevsky2012imagenet}. Since AlexNet specializes in object classifications (won 1st position for ImageNet competition in 2012 with dataset containing 15 million of images and with 1000 of different object classes), a few adjustments are made in the network layers to make architecture capable of doing the segmentation task. Earlier, network's convolution layers are maintained for as usual features extraction, but, the fully connected (FC) layers are removed and replaced with equivalent convolution layers as FC layers throw away the spatial information which is crucial for the segmentation task. 

\paragraph{\textbf{FCN-VGGNet}}

In 2014, K. Simonyan and A. Zisserman of the Visual Geometry Group (VGG) from the University of Oxford proposed a CNN model known as \textit{VGG-16} \cite{simonyan2014very} and secured the first and second place in localization and classification, respectively in the ImageNet Challenge 2014, with a test error rate of 7.3\% and test accuracy of 92.7\% in ILSVRC 2014 submission. VGG-16 architecture is customized to three models, \textit{FCN-32s}, \textit{FCN-16s} and \textit{FCN-8s}, where each model magnifies the output with different upsampling layers \cite{shelhamer2017fully}. For FCN-32s, the fully connected layers are convolutionized and a 32-pixel stride size is applied in deconvolution layers. This works well for object classification, localization and detection tasks as precise pixel-wise predictions are not the main priority. FCN-16s and FCN-8s perform better for object segmentation tasks, as both extract extra low-level features from the input image to produce a more precise output result by 16 x 16 and 8 x 8 pixel blocks, respectively. 

In FCN-16s, the final output is a product of upsampling of two layers, which are upsampled from the fourth pooling layer and (upsampling of the seventh convolutional layer) x 2. For FCN-8s, the final output of the model is a product of upsampling of the third pooling layer, (upsampling of the fourth pooling layer) x 2 and (upsampling of the seventh convolutional layer) x 4.

\paragraph{\textbf{PSPNet}}

In 2016, \textit{Zhou et al.} proposed a network architecture called Pyramid Scene Parsing Network (\textit{PSPNet}) \cite{zhao2016pyramid}. In this work, the research team from Chinese University of Hong Kong and SenseTime Group Limited used the global context information proficiency by aggregation of a different-region-based context. The global spatial context is important since it provides suggestions on the distribution of the segmentation classes and therefore, tackling issues related to mismatched relationship, confusion categories and inconspicuous classes. This procedure is done by the combination of a pyramid pooling module (applying four different sizes of kernel pooling layers) and proved to have an effective global prior representation capable to produce good results in the scene parsing tasks and pixel-level predictions. Other key features of this architecture include: integrating dilated convolutions as a means of modifying the base architecture of Residual Network \textit{ResNet} \cite{he2016deep}, a 152-layer network architecture proposed by Microsoft Research Asia in late 2015. The proposed method achieved the first place in ImageNet scene parsing challenge 2016, PASCAL VOC 2012 benchmark and Cityscapes benchmark. The paper also reported that PSPNet model yields an accuracy of 85.4\% (by mean intersection over union (mIoU)) on PASCAL VOC 2012 and accuracy of 80.2\% on Cityscapes. 

\subsubsection{Transfer Learning}

\noindent Having a sufficient amount of data is crucial to effectively train a deep neural network. However, in reality especially for medical imaging, it is not often possible to have large datasets and this is why the idea of transfer learning has emerged in recent years to produce significant results for smaller medical imaging datasets. Instead of training the deep learning methods from scratch, transfer learning is a method that uses the weights from pre-trained models and then, fine-tuning the network with the new dataset. 

\begin{figure}[!h] 
	\centering
	\includegraphics[width=\textwidth]{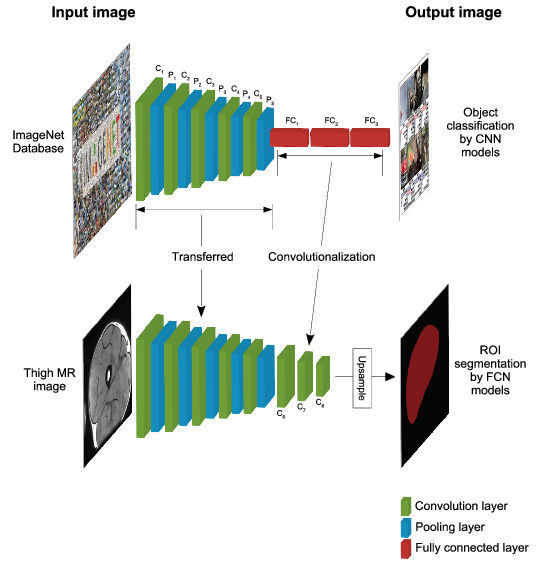}
	\captionsetup{justification=justified,singlelinecheck=false}
	\caption{Transfer learning procedure of deep CNNs to obtain optimized weights initializations. Three fully connected layers of  CNN were removed and replaced by three convolutional layers, making the pre-trained model fully convolutional.}
	\label{Fig8}
\end{figure}

In this work, the weights associated to these pre-trained models are subsequently used to fine-tune the deep learning methods for training the thigh MRI datasets. This transfer learning process is demonstrated in Figure \ref{Fig8}.

\paragraph{\textbf{Configuration of GPU Machine for Experiments}}

(1) Hardware: CPU - Intel i7-6700 @ 4.00Ghz, GPU - NVIDIA 1080 TITAN X 11Gb, RAM - 32Gb DDR5 (2) Software: Caffe \cite{jia2014caffe}.



\section{Results and Discussion}

\noindent Two datasets including a total of 1110 MR images were manually labelled by the domain experts. These datasets of MRI are then trained, validated and tested by the state-of-the-art semantic segmentation models that are FCN-AlexNet, FCN-VGG-16 (consists of FCN-32s, FCN-16s and FCN-8s) and PSPNet. Both mid-scan (MD) and whole-scan (WD) MRI datasets were split into the configuration of 70\%, 10\% and 20\% for training, validation and testing, respectively (MD: 77, 11, 22 images and WD: 700, 100, 200 images), with 5-fold cross-validation procedure (5 processing batches to analyse all MR images). These models were trained on deep learning library called Caffe \cite{jia2014caffe} with solver type is stochastic gradient descent \cite{bottou2012stochastic}, the learning rate of 0.0001 and number of epochs are set to 50.

To evaluate the performance of these FCN models, all of the semantically segmented (tested) images were compared and measured with the corresponding labelled (manually segmented ROI) images. In addition to \textit{Jaccard Similarity Index (JSI)}, \textit{Sensitivity}, \textit{Specificity}, \textit{Matthews Correlation Coefficient (MCC)} and \textit{Dice Similarity Coefficient (DSC)} indexes were also employed as the performance metrics for the evaluation of segmented ROI. \textit{MCC} \cite{matthews1975comparison} is defined as:
\begin{equation}
MCC=\frac { TP \times TN-FP \times FN }{ \sqrt { (TP+FP)(TP+FN)(TN+FP)(TN+FN) }  }   ,
\label{Eq6}
\end{equation}
where \textit{TP}, \textit{TN}, \textit{FP} and \textit{FN} denote True Positives, True Negatives, False Positives and False Negatives, respectively.

Table \ref{Table1} summarizes the mean index of semantic segmentation results of all the deep learning models on mid-scan (MD) and whole-scan (WD) datasets.

\begin{table}[h]
\centering
\captionsetup{justification=justified,singlelinecheck=false}
\caption{Mean performance index for all models on mid-scan (MD) and whole-scan (WD) datasets.}
\label{Table1}
\resizebox{\textwidth}{!}{\begin{tabular}{c|c|c|c|c|c|c|c|c|c|c|}
\cline{2-11}
                                  & \multicolumn{2}{c|}{JSI}          & \multicolumn{2}{c|}{DSC}          & \multicolumn{2}{c|}{Sensitivity}  & \multicolumn{2}{c|}{Specificity}  & \multicolumn{2}{c|}{MCC}          \\ \hline
\multicolumn{1}{|c|}{Network}     & MD              & WD              & MD              & WD              & MD              & WD              & MD              & WD              & MD              & WD              \\ \hline
\multicolumn{1}{|c|}{PSPNet}  & 0.8263          & 0.9194          & 0.9026          & 0.9565          & 0.9029          & 0.9583          & 0.9252          & 0.9611          & 0.828           & 0.9209          \\ \hline
\multicolumn{1}{|c|}{FCN-AlexNet} & 0.8487          & 0.8923          & 0.9169          & 0.9332          & 0.901           & 0.9446          & \textbf{0.9804} & \textbf{0.9769} & 0.8925          & 0.9153          \\ \hline
\multicolumn{1}{|c|}{FCN-32s}     & 0.8198          & 0.9211          & 0.899           & 0.9581          & 0.8796          & 0.9768          & 0.9776          & 0.9467          & 0.8712          & 0.9232          \\ \hline
\multicolumn{1}{|c|}{FCN-16s}     & 0.8669          & 0.9199          & 0.9266          & 0.9574          & 0.9399          & 0.9782          & 0.9743          & 0.9442          & 0.9059          & 0.922           \\ \hline
\multicolumn{1}{|c|}{FCN-8s}      & \textbf{0.8973} & \textbf{0.9316} & \textbf{0.9442} & \textbf{0.9639} & \textbf{0.9708} & \textbf{0.9822} & 0.9748          & 0.9514          & \textbf{0.9282} & \textbf{0.9336} \\ \hline
\end{tabular}}
\end{table}

Overall, FCN-8s performed best in regards to the semantic segmentation task of the ROI of thigh MR images, with an average accuracy of 0.8973 (89.73\%) for MD and 0.9316 (93.16\%) for WD by \textit{JSI}. The average \textit{DSC} performance index for the corresponding datasets was also notably higher than \textit{JSI}, i.e. 0.9442 (94.42\%) and 0.9639 (96.39\%), for MD and WD, respectively. Due to the less-forgiving nature of its measurement, \textit{JSI} will be used comprehensively from here after. \textit{DSC} was included partially because it was universally employed as a means of performance evaluation in machine learning. Although both \textit{JSI} and \textit{DSC} adapted a slightly different formulation from one another, both indexes are good for performance assessment of segmentation task. 

FCN-8s excelled throughout the performance indices used as it was ranked first in \textit{JSI}, \textit{DSC}, \textit{Sensitivity} and \textit{MCC} and ranked third in the \textit{Specificity} performance index. However, the difference in \textit{Specificity} was not substantial from the first-ranked FCN-AlexNet model (differences of 0.0056 and 0.0255 for MD and WD, respectively).

\subsection{FCN-8s model with different solver types for MD and WD}

\noindent Stochastic Gradient Descent (SGD) \cite{bottou2012stochastic} is arguably predominantly used to train deep learning models, mainly due to its strength of simplicity in implementation and fast processing, even for problems (or datasets) with many training patterns. Semantic segmentation results in Table \ref{Table1} also generated by the application of SGD as its solver type in all pre-trained models. 

Table \ref{Table2} demonstrates the performance impact on the application of different solver methods used for semantic segmentation in thigh MR datasets by FCN-8s model.

\begin{table}[h]
\centering
\captionsetup{justification=justified,singlelinecheck=false}
\caption{Mean \textit{JSI} performance index for different solver types used in FCN-8s on mid-scan (MD) and whole-scan (WD) datasets.}
\label{Table2}
\resizebox{\textwidth}{!}{\begin{tabular}{c|c|c|c|c|c|c|c|}
\cline{2-8}
                          & \multicolumn{5}{c|}{MD}                               & \multicolumn{2}{c|}{WD}  \\ \cline{2-8}  
                          & AdaGrad \cite{duchi2011adaptive} & Adam \cite{kingma2014adam}            & NAG \cite{nesterov1983method}    & RMSprop \cite{hinton2012rmsprop} & SGD    & Adam            & SGD    \\ \hline
\multicolumn{1}{|c|}{JSI} & 0.8778  & \textbf{0.927} & 0.8972 & 0.8829  & 0.8973 & \textbf{0.9469} & 0.9316 \\ \hline
\end{tabular}}
\end{table}

From Table \ref{Table2}, the various applications of solver types on FCN-8s during the training and validation suggested that Adam solver generally performs better for the semantic segmentation of ROI in thigh MRI, with both MD and WD datasets having a \textit{JSI} performance increment of 0.0297 (2.97\%) and 0.0153 (1.53\%) in terms of accuracy, respectively compared to FCN-8s with SGD solver. Unfortunately, none of the results can be generated with AdaDelta solver \cite{zeiler2012adadelta} on FCN-8s due to the lower percentage ($\approx{70\%}$) of learning accuracy and higher loss during the training, even after the number of epoch was increased.

\subsection{Segmentation of MRI thigh muscle on All-scan Dataset}

\noindent The main intention of distributing the dataset into two (MD and WD) is to observe for any significance in \textit{JSI} performance index for the semantic segmentation task of the ROI of the thigh MRI due to the substantial differences of the total number of images in these datasets (110 and 1000 images for MD and WD, respectively). As it was proven by the results in the previous sections, the performance of each model is marginally improved with the increased number of images in the WD dataset. 
 
We further investigate the performance of these models by combining the MD and WD to form an all-scan dataset (AD). For this purpose, 1000 images of WD are used for training and validation, whereas 110 images of MD are used for testing. The new training, validation and testing configurations for AD are then segregated as 900, 100 and 110 images, respectively. The hypothesis of forming this dataset is that the testing performances for each model should be improved as more images are incorporated in training the models, compared to MD and WD datasets. Table \ref{Table3} shows the results of this procedure, with results of both MD and WD are also included, for comparisons.

\begin{table}[h]
\centering
\captionsetup{justification=justified,singlelinecheck=false}
\caption{Mean \textit{JSI} performance index of all models on mid-scan (MD), whole-scan (WD) and all-scan (AD) datasets.}
\label{Table3}
\resizebox{\textwidth}{!}{\begin{tabular}{cc|c|c|c|c|c|c|c|}
\cline{3-9}
                              &    & \multicolumn{7}{c|}{Model}                                                                        \\ \cline{3-9} 
                              &    & \multicolumn{2}{c|}{FCN-AlexNet} & \multicolumn{2}{c|}{FCN-8s}       & FCN-16s & FCN-32s & PSPNet \\ \cline{3-9} 
                              &    & Adam            & SGD            & Adam            & SGD             & SGD     & SGD     & SGD    \\ \hline
\multicolumn{1}{|c|}{}        & MD & 0.897          & 0.8487         & \textbf{0.927} & 0.8973          & 0.8669  & 0.8198  & 0.8263 \\ \cline{2-9} 
\multicolumn{1}{|c|}{Dataset} & WD & 0.9253          & 0.8923         & \textbf{0.9469} & 0.9316          & 0.9199  & 0.9211  & 0.9194 \\ \cline{2-9} 
\multicolumn{1}{|c|}{}        & AD & 0.9228          & 0.8988         & 0.9388          & 0.9391 & 0.9378  & \textbf{0.9412}  & 0.9081 \\ \hline
\end{tabular}}
\end{table}

As expected, overall FCN-8s (with Adam solver) performed better compared to the other models, throughout all the datasets (MD, WD or AD), with the performance of FCN-8s for WD dataset (94.69\%) remains the best. However, surprisingly, the incorporation of SGD solver on FCN-32s performed better (compared to the other models within AD), with average \textit{JSI} performance of 0.9412 (94.12\%), an increment of 0.201 (or 2.01\%) by \textit{JSI} compared to FCN-32s for WD. Also note that the mean \textit{JSI} performances for FCN-8s (with both Adam and SGD solver methods), FCN-16s and FCN-32s, in AD dataset are also close to one another, with only a small performance difference between them ($\pm{0.0034}$ or 0.34\%). This suggests that earlier hypothesis is inaccurate for the most of the deep learning models employed, as overall performance improved only for FCN-32s. The main reason for failed hypothesis is more divergent testing data that consists of single scan of 110 individual subjects for training data that consists of sequential scans of only 50 individual subjects in this dataset.

The average processing time to inference a single MRI of thigh muscle also stands as an important performance indication for the means of evaluation of all models. Table \ref{Table4} depicts the average processing (semantic segmentation testing) time (in seconds) for a single thigh MR image for all proposed models. 

\begin{table}[h]
\centering
\captionsetup{justification=justified,singlelinecheck=false}
\caption{Mean processing (testing) time (in seconds) for all models to semantically segment an ROI of thigh MR image.}
\label{Table4}
\resizebox{\textwidth}{!}{\begin{tabular}{cc|c|c|c|c|c|c|c|c|c|c|c|c|}
\cline{3-14}
                              &    & \multicolumn{12}{c|}{Model}                                                                                                                                      \\ \cline{3-14} 
                              &    & \multicolumn{2}{c|}{FCN-AlexNet} & \multicolumn{5}{c|}{FCN-8s}                     & \multicolumn{2}{c|}{FCN-16s} & \multicolumn{2}{c|}{FCN-32s} & PSPNet        \\ \cline{3-14} 
                              &    & Adam            & SGD            & AdaGrad & Adam & NAG  & RMSprop & SGD           & Adam          & SGD          & Adam          & SGD          & SGD           \\ \hline
\multicolumn{1}{|c|}{}        & MD & 0.34            & 0.24           & 0.46    & 0.64 & 0.45 & 0.37    & 0.61          & 0.58          & 0.39         & 0.61          & 0.55         & \textbf{0.05} \\ \cline{2-14} 
\multicolumn{1}{|c|}{Dataset} & WD & 0.18            & 0.15           & -       & 0.1  & -    & -       & \textbf{0.03} & -             & 0.03         & -             & 0.03         & 0.05          \\ \cline{2-14} 
\multicolumn{1}{|c|}{}        & AD & 0.1             & 0.09           & -       & 0.17 & -    & -       & \textbf{0.03} & -             & 0.04         & -             & 0.03         & 0.05          \\ \hline
\end{tabular}}
\end{table}

FCN-8s with SGD solver (in AD and WD) was the best in terms of computation time for the semantic segmentation task of thigh MRI, with 30 ms (0.03 sec) average processing time per image. That was 100 times faster than the proposed method in \cite{ahmad2014enhancement} (3 sec), 567 times faster than the proposed method in \cite{ahmad2014atlas} (17 sec) and 733 times faster the proposed method in \cite{ahmad2018automatic}. In correlation to the best performer in terms of accuracy (FCN-8s with Adam solver in WD; average processing time of 0.1 sec per image), the processing time was 30, 170 and 220 times faster than the frameworks in \cite{ahmad2014enhancement}, \cite{ahmad2014atlas} and \cite{ahmad2018automatic}, respectively. This indicates that in general, the FCN-8s model is ideal for the domain of semantic segmentation as it is not only accurate, but also considerably fast. Overall, the application of any pre-trained model for testing (semantic segmentation) the ROI of thigh MRI is reliable in terms of processing time, with FCN-8s (with Adam solver in MD) taking the longest time - 0.64 sec. And that still is 4.7 times faster than the proposed method in \cite{ahmad2014enhancement}.

This was also confirmed: 1) the general influence of utilizing the simplicity of update value formulation of the weight ${W}_{t+1}$ \cite{bottou2012stochastic} for SGD, evidently results in faster computation, compared to the complexity of Adams's weight update \cite{kingma2014adam}; and 2) vice-versa for the accuracy department, as Adam generally performed better throughout all datasets, compared to SGD.

\subsection{Post-processing}
\begin{figure}[!h] 
\centering
\includegraphics[width=\textwidth]{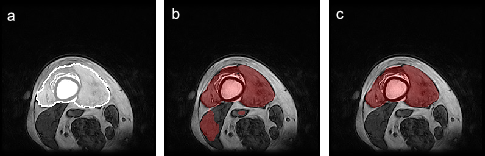}
\captionsetup{justification=justified,singlelinecheck=false}
\caption{Image filtering and morphological to the semantic segmentation image. (a) Ground truth label (in white) with original MR image. (b) Semantic segmentation result by FCN8-s model (for this sample: with Adam solver in WD dataset). (c) Post-processing result of (b).}
\label{Fig9}
\end{figure}

\noindent Notwithstanding the promising results set out above, there is still a need to investigate the possibility of improving the end results of semantic segmentation by the FCN-8s model. Here, the inclusion of a post-processing stage by image processing techniques will be explored, with the goal of enhancing the ROI segmentation result. This FCN-8s model dedicated experimentations will be applied to both Adam and SGD solvers on all datasets (MD, WD and AD). The process involves image filtering and morphological procedures to the segmented ROI. Figure \ref{Fig9} demonstrates the significance of the inclusion of this stage and Table \ref{Table5} shows the mean \textit{JSI} performance index with the implementation of post-processing stage by image processing techniques described earlier.

\begin{table}[h]
\centering
\captionsetup{justification=justified,singlelinecheck=false}
\caption{Mean \textit{JSI} performance index of FCN-8s model on mid-scan (MD), whole-scan (WD) and all-scan (AD) datasets, without and with post-processing, respectively.}
\label{Table5}
\begin{tabular}{cc|c|c|c|c|}
\cline{3-6}
                              &    & \multicolumn{4}{c|}{FCN-8s}                                                              \\ \cline{3-6} 
                              &    & \multicolumn{2}{c|}{Without post-processing} & \multicolumn{2}{c|}{With post-processing} \\ \cline{3-6} 
                              &    & Adam                  & SGD                  & Adam                 & SGD                \\ \hline
\multicolumn{1}{|c|}{}        & MD & 0.927                & 0.8973               & 0.927                    & 0.9006                  \\ \cline{2-6} 
\multicolumn{1}{|c|}{Dataset} & WD & 0.9469                & 0.9316               & \textbf{0.9502}                    & 0.9325                  \\ \cline{2-6} 
\multicolumn{1}{|c|}{}        & AD & 0.9388                & 0.9391               & 0.9398                    & 0.9392                  \\ \hline
\end{tabular}
\end{table}

The results above demonstrated that by incorporating the proposed post-processing stage, the semantic segmentation output can be enhanced. The overall accuracy performances are increased throughout all of the datasets, with FCN-8s (Adam solver in WD) achieved the best result. However, the performance increment was minimal, due to the fact that only $<1\%$ (or 0.86\% to be exact across output in Table \ref{Table5}) of semantically segmented images are affected by this over-segmentation condition as shown in Figure \ref{Fig9}(b). With additional processing, the overall mean processing time to process a thigh MR image also increased. It is measured that an extra 17 ms (0.017 sec) is needed for this purpose (for FCN-8s with Adam solver in WD), increasing the overall processing time to $0.1+0.017=0.117$ sec.

\subsection{Comparison with benchmark algorithms}

The closest state-of-the-art segmentation methods using the similar definition of ROIs were done by \textit{Ahmad et. al.} \cite{ahmad2014atlas, ahmad2014enhancement, ahmad2018automatic}. They first established a semi-automated segmentation method for mid-scan ROI segmentation \cite{ahmad2014atlas} and achieved a mean JSI of 0.95 (or 95\%). The work was further investigated towards the automatic segmentation of sequential MR images in a dataset (of a single subject) \cite{ahmad2014enhancement} by the combination frameworks of the earlier method in \cite{ahmad2014atlas} and image processing techniques and achieved an average JSI of 0.9334 (or 93.34\%). Although both frameworks yield reliable and accurate segmentation results, these frameworks still need minimal supervision to precisely delineate the muscle borders. A fully automated segmentation method was later developed in \cite{ahmad2018automatic} by the integration of statistical analysis, image processing and computer vision algorithms, which requires no interactions from the user and offers good performance in terms of segmentation accuracy (mean JSI of 0.8526) of the ROI (quadriceps, femur and marrow) and processing time.

This paper has demonstrated that FCN-8s with post-processing has out-performed the benchmark algorithms and therefore warrants the in-depth analysis before further application. In addition, in terms of processing time, it was 26, 145 and 188 times faster than the frameworks in \cite{ahmad2014enhancement}, \cite{ahmad2014atlas} and \cite{ahmad2018automatic}, respectively.

\section{Conclusion}

\noindent In this work, five deep learning models FCN-AlexNet, FCN-32s, FCN-16s, FCN-8s and PSPNet have been explored for the semantic segmentation task of the ROI which include quadriceps, bone, and marrow on MRI of thigh muscles. FCN-8s emerges as the best all-around deep learning model for the task, and Adam solver type works the best overall for the thigh MR images, as compared to the other solver types. The FCN-8s with Adam solver, trained, validated and tested on mid-scan (MD), whole-scan (WD) and all-scan (AD) datasets, produced a mean semantic segmentation accuracy by \textit{JSI} of 0.927 (92.7\%), 0.9469 (94.69\%) and 0.9388 (93.88\%), within an average processing time per image of 0.64 sec, 0.1 sec and 0.17 sec, respectively for the corresponding datasets. 

With the implementation of post-processing stage of image filtering and morphology to the ROI segmented output, the results are enhanced to 0.927 (92.7\%), 0.9502 (95.02\%) and 0.9398 (93.98\%) for MD, WD and AD, respectively. The best \textit{JSI} performance result of 0.9502 (95.02\%) suggests that FCN-8s model with Adam solver in WD has out-performed the mean \textit{JSI} benchmark of 0.95 (or 95\%) set in \cite{ahmad2014atlas} for the semi-automatic segmentation framework. However, this semi-automatic segmentation framework suggested in \cite{ahmad2014atlas} requires manual tuning of image processing parameters for the individual cases which can be very tedious work especially for domain experts. Whereas, the deep learning models present end-to-end solution with quick processing time for inferencing. The output performance of FCN-8s (without the inclusion of post-processing stage) i.e. 0.947 (or 94.7 \%) is comparable to the previous benchmark achieved. 

In the future work: 1) to explore other potential deep learning architectures that are used for the semantic segmentation task (such as U-Net \cite{ronneberger2015u}, SegNet \cite{badrinarayanan2015segnet} and FCN-DenseNet \cite{jegou2016one}) and investigate, evaluate and compare their performances with the current benchmark method; and 2) the prospect of developing novel deep learning architecture that specifies and optimizes to the ROI segmentation with post-processing technique included in it.


\bibliography{ref_ult}

\end{document}